\documentclass[10pt,twocolumn,letterpaper]{article}

\usepackage{float}
\usepackage{stfloats}
\usepackage{makecell}
\usepackage{multirow}
\usepackage{multicol}
\usepackage{hhline}
\usepackage{listings}
\usepackage{graphicx}
\newcommand{\mycc}{\cellcolor{lightgray}}

\usepackage{cvpr}      

\definecolor{cvprblue}{rgb}{0.21,0.49,0.74}
\usepackage[pagebackref,breaklinks,colorlinks,allcolors=cvprblue]{hyperref}

\def\paperID{14784} 
\def\confName{CVPR}
\def\confYear{2025}

\title{GLane3D : Detecting Lanes with Graph of 3D Keypoints}

\author{Halil İbrahim Öztürk \quad Muhammet Esat Kalfaoğlu \quad Ozsel Kilinc\\
Togg/Trutek AI Team\\
{\tt\small \{ibrahim.ozturk,esat.kalfaoglu,ozsel.kilinc\}@togg.com.tr}
}

\begin{document}
\maketitle
\begin{abstract}
Accurate and efficient lane detection in 3D space is essential for autonomous driving systems, where robust generalization is the foremost requirement for 3D lane detection algorithms. Considering the extensive variation in lane structures worldwide, achieving high generalization capacity is particularly challenging, as algorithms must accurately identify a wide variety of lane patterns worldwide. Traditional top-down approaches rely heavily on learning lane characteristics from training datasets, often struggling with lanes exhibiting previously unseen attributes. To address this generalization limitation, we propose a method that detects keypoints of lanes and subsequently predicts sequential connections between them to construct complete 3D lanes. Each key point is essential for maintaining lane continuity, and we predict multiple proposals per keypoint by allowing adjacent grids to predict the same keypoint using an offset mechanism. PointNMS is employed to eliminate overlapping proposal keypoints, reducing redundancy in the estimated BEV graph and minimizing computational overhead from connection estimations. Our model surpasses previous state-of-the-art methods on both the Apollo and OpenLane datasets, demonstrating superior F1 scores and a strong generalization capacity when models trained on OpenLane are evaluated on the Apollo dataset, compared to prior approaches.
    \vspace*{-0.5cm}

\end{abstract}
    
\section{Introduction}
\label{sec:intro}
    \vspace*{-0.2cm}

Robust 3D lane detection is crucial for various autonomous driving functionalities, such as lane keeping, lane departure warning, and trajectory planning. However, lane detection methods often struggle in complex scenarios. While some systems utilize LiDAR or multi-sensor setups, camera-only setups are increasingly preferred due to their cost-effectiveness for lane marking detection. A major challenge remains the accurate detection of 3D lane boundaries, which is essential for safe navigation and provides critical information for path planning and vehicle control.

Historically, lane detection primarily relied on 2D methods, utilizing techniques such as segmentation-based \cite{neven2018towards, pan2018spatial, zheng2021resa, ghafoorian2018gan, hou2019learning, pizzati2020lane, zou2019robust, lee2017vpgnet, xu2020curvelane}, anchor-based \cite{li2019line, tabelini2021keep, zheng2022clrnet, su2021structure, wang2018lanenet, jin2022eigenlanes}, or key-point-based approaches \cite{qu2021focus, ko2021key, wang2022keypoint, huval2015empirical, xu2022rclane}. To extend these 2D detections into a 3D space, inverse projection methods were applied; however, this approach often fell short, as the lack of depth information limited accuracy in representing the true 3D structure of lanes. Another workaround \cite{yan2022once} involved performing a separate depth estimation before projection, but this demanded exceptionally accurate depth data and flawless 2D segmentation. Recently, end-to-end methods for 3D lane detection \cite{chen2022persformer, luo2023latr, pittner2024lanecpp, guo2020gen}, which leverage front-view camera images, have emerged as a promising solution. These methods often project perspective-view features into a bird’s-eye view (BEV) using inverse perspective mapping (IPM) or lift-splat-shoot (LSS) methods. Alternatively, some models utilizes 3D lane anchors into the perspective view, predict necessary offsets, and re-project the results back.

Lane detection in both 2D and 3D spaces is primarily approached in two ways: instance-based (top-down) methods \cite{tabelini2021laneatt, chen2022persformer, pittner2024lanecpp, luo2023latr, li2019linecnn, zheng2022clrnet}, which directly predict entire lane instances, and bottom-up methods, which detect individual lane components, such as keypoints, and later assemble these into full lane lines through post-processing steps \cite{wang2023bev, qu2021focus, wang2022keypoint}. Segmenting pixels as lane or background is generally more straightforward than predicting entire lane instances with complex parameters. This distinction enhances the generalization capabilities of bottom-up methods, enabling them to detect lane parts even in previously unseen scenarios, unlike top-down approaches, which may struggle with novel lane types. On the other hand, bottom-up techniques face challenges in grouping detected keypoints to form cohesive lanes. For instance, \cite{wang2023bev} employs clustering-based post-processing from learned keypoint features, while \cite{wang2022keypoint} determines the direction each keypoint points to a common lane start, allowing for grouping, and \cite{qu2021focus} uses iterative association of keypoints to progressively construct lanes.

To simplify keypoint grouping in the post-processing stage, differently from the other bottom-up approaches, we formulate lane detection as locating keypoints which are part of the predicted directed graph, and predicting connections between consecutive ones, allowing lane extraction using shortest path algorithms between initial and final keypoints. Detecting each keypoint of a lane accurately is crucial, as missing keypoints can fragment lanes. To address this, we propose using multiple keypoints within a $d_x$ distance for each target keypoint along the lane and apply small offsets to refine their positions. While introducing redundant keypoints decreases the likelihood of missed detections, it also increases computational cost. To manage this trade-off, we use a PointNMS operation to retain only the strongest keypoints by removing redundancies.
\begin{figure*}[t]
  \centering
  \includegraphics[width=\textwidth]{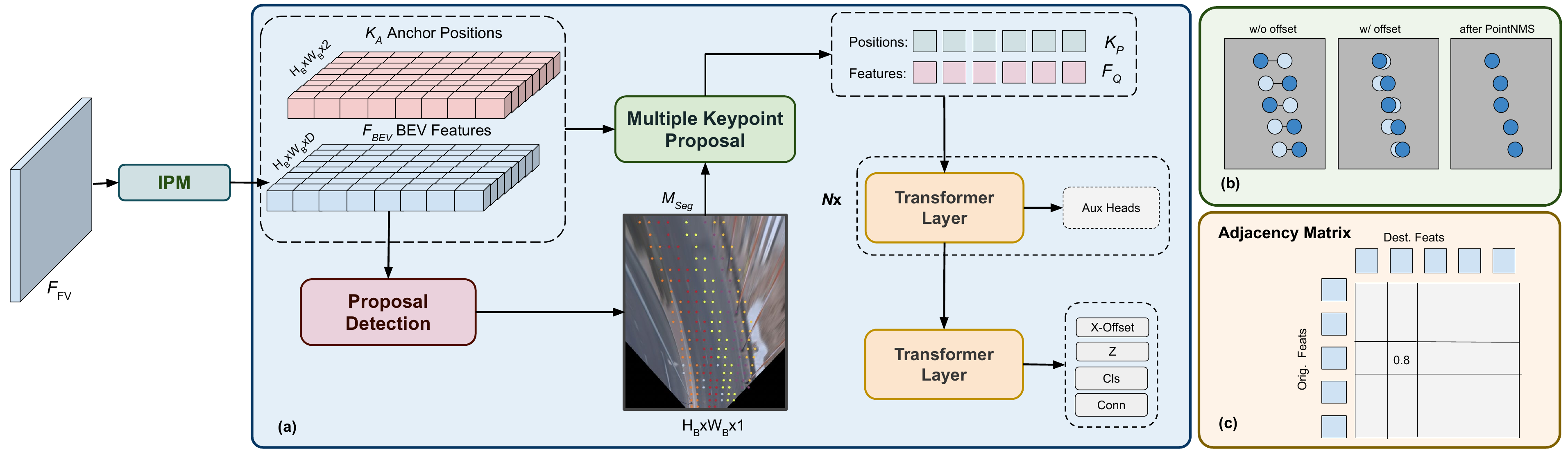}
  \caption{\textbf{The overall architecture.} GLane3D is a keypoint-based 3D lane detection method. A directed graph of 3D keypoints is used to generate lane instances. In part (a), multiple proposals improve the recall of keypoint detection, while in part (b), PointNMS selects the strongest proposals to reduce ambiguity in the directed graph. The estimated adjacency matrix, shown in part (c), enables extraction of directed connections.}
  \label{fig:overall}
    \vspace*{-0.2cm}
\end{figure*}

Our main contributions are the following:
    \vspace*{-0.5cm}

\begin{itemize}
    \item We propose \textbf{GLane3D}, a keypoint-based 3D lane detection method. It uses directed connection estimation between keypoints for efficient lane extraction. By generating multiple keypoint proposals, we improve detection recall. The PointNMS algorithm selects the strongest proposal, reducing ambiguity and computational overhead.
    \item Using Inverse Perspective Mapping (IPM) with customized BEV locations to compute sampling points at the Frontal View (FV), instead of using uniformly distributed BEV locations, reduces sparsity in regions near the ego vehicle and alleviates saturation in distant areas. 
    \item Thanks to its innovative design, GLane3D exhibits high generalization ability. We evaluate this through cross-dataset testing, where the OpenLane\cite{chen2022persformer}-trained model performs well on the Apollo\cite{guo2020gen} dataset, demonstrating GLane3D’s strong generalization compared to previous methods.
    \item  The camera-only GLane3D outperforms state-of-the-art methods on both OpenLane\cite{chen2022persformer} and Apollo\cite{guo2020gen} datasets, surpassing previous methods across all OpenLane test categories. GLane3D with Camera+Lidar fusion also leads in F1 score. It also achieves the highest frames per second (FPS) compared to other models.
    \vspace*{-0.3cm}
\end{itemize}

\section{Related Works}
    \vspace*{-0.1cm}
\label{sec:formatting}

\subsection{2D Lane Detection}

Despite promising advancements in 2D lane detection, a notable gap remains between 2D results and the precise 3D positions required for real-world applications. 2D lane detection methods can be broadly categorized into four main approaches. \textbf{Segmentation-based} approaches \cite{neven2018towards, pan2018spatial, zheng2021resa, ghafoorian2018gan, hou2019learning, pizzati2020lane, zou2019robust, lee2017vpgnet, xu2020curvelane} segment lane-line portions from perspective images, followed by post-processing to obtain lane instances. \textbf{Anchor-based} methods \cite{li2019line, tabelini2021keep, zheng2022clrnet, su2021structure, wang2018lanenet, jin2022eigenlanes} use lane-line anchors (e.g., predefined points as line anchors) to regress relative offsets to targets. Another anchor type, row-based anchors \cite{yoo2020end, qin2020ultra, liu2021condlanenet, qin2022ultra}, classifies row-wise pixels into lanes. \textbf{Curve-based} approaches \cite{van2019end, tabelini2021polylanenet, liu2021end, feng2022rethinking, lu2021super, wang2020polynomial} predict polynomial parameters to fit a curve over lanes, leveraging prior lane knowledge. \textbf{Keypoint-based} approaches \cite{qu2021focus, ko2021key, wang2022keypoint, huval2015empirical, xu2022rclane} flexibly model lanes by first estimating point locations and then grouping them. For grouping, some methods use global point prediction \cite{wang2022keypoint} or a combination of global and local predictions \cite{xu2022rclane}, while others predict neighboring points based on geometry \cite{qu2021focus}. These methods often involve an extra step, such as heatmap extraction, to match keypoints effectively.

\subsection{3D Lane Detection}

To obtain accurate 3D lane positions, 3D lane detection methods estimate from a front-view (FV) camera. Projection of FV features into Bird's Eye View (BEV) space enables 3D lane detection as seen in \cite{chen2022persformer, guo2020gen, pittner2024lanecpp, li2022reconstruct, pittner20233d}. Projections for 3D lane detection can be categorized into two main types: Lift-Splat-Shoot (LSS)-based \cite{pittner2024lanecpp, luo2022m} and Inverse Perspective Mapping \cite{mallot1991inverse} (IPM)-based \cite{chen2022persformer, guo2020gen, li2022reconstruct}. LSS projections leverage depth estimation learned directly from the front-view camera, where depth estimation is supervised in some works \cite{pittner2024lanecpp, luo2022m}. Other methods, estimate 3D locations from FV directly without feature projection \cite{luo2024dv, luo2023latr}  by predicting 3D lane points via lane and point queries similar to PETR \cite{liu2022petr}. Additionally, anchor lane-lines projected into 3D space can regress offsets without feature projection \cite{huang2023anchor3dlane}. Methods that implicitly learn 3D projection instead of explicit FV-to-BEV transformation struggle with camera location changes. 3D lane estimates can also be obtained by combining 2D lane predictions with dense depth estimation \cite{yan2022once} or by directly projecting 2D predictions into 3D space \cite{guo2020gen}.

FV to BEV projection enables leveraging 2D methods in 3D lane detection. \textbf{Anchor-based} approaches \cite{chen2022persformer, luo2022m, li2022reconstruct} regress lateral offsets from anchor lane-lines to obtain target lanes, while \textbf{curve-based} approaches \cite{bai2023curveformer, pittner20233d, pittner2024lanecpp, kalfaoglu2024topobda} predict curve parameters. LaneCPP \cite{pittner2024lanecpp}, uses physical priors to regulate learned parameters. The \textbf{keypoint-based} approach BEVLaneDet \cite{wang2023bev} uses anchor keypoints centered in BEV grids, the keypoints are regressed to target lane, followed by grouping keypoints step with learned embeddings.

Viewing lane detection as multi-point object detection limits generalization, as models may struggle with unseen lane configurations. Keypoint-based methods, treating lane detection as detecting and grouping local parts into line instances, enhance generalization. Keypoint-based methods in 2D or 3D handle grouping with either clustering-based or local geometric offset predictions, or both. GLane3D, estimates connections between consecutive keypoints and applies PointNMS to reduce redundancy, facilitating efficient grouping. 

    \vspace*{-0.1cm}

\section{Methodology}

GLane3D detects 3D keypoints \( \mathbf{k}_i \in \mathbf{K} \), where \( \mathbf{k}_i = (x_i, y_i, z_i) \) for \( i = 1, 2, \dots, S \), and predicts an adjacency matrix \( \mathbf{A} \in \mathbb{R}^{S \times S} \), where \( \mathbf{A}[i][j] \) represents the probability of a connection between \( \mathbf{k}_i \) and \( \mathbf{k}_j \). We construct a directed graph \( G \) from the set \( C \) of directed connections, extracted with Eq. \ref{eq:con_from_adj}, and the estimated set \( \mathbf{K} \) consists of the predicted keypoints. Lane instances are extracted through a straightforward process from the directed graph \( G \).
\begin{equation}
     C = \{ (\mathbf{k}_i, \mathbf{k}_j) \mid \mathbf{A}[i][j] > t_a \}.
     \label{eq:con_from_adj}
\end{equation}
The predictions are based on features projected from the frontal view (FV) to the bird’s-eye view (BEV), denoted as \( \mathbf{F}_{BEV} \in \mathbb{R}^{c \times H_b \times W_b} \), which is obtained by applying the inverse perspective mapping (IPM) projection to the FV features \( \mathbf{F}_{FV} \), extracted by the backbone network from the input image \( \mathbf{I} \in \mathbb{R}^{3 \times H \times W} \).

\subsection{Model Overview}

\begin{figure}
  \centering
    \begin{subfigure}{0.0\linewidth}
            \phantomsubcaption
        \label{fig:stages_a}
    \end{subfigure}
        \begin{subfigure}{0.0\linewidth}
        \phantomsubcaption
        \label{fig:stages_b}

    \end{subfigure}
        \begin{subfigure}{0.0\linewidth}
        \phantomsubcaption

        \label{fig:stages_c}
    \end{subfigure}
    
  \begin{subfigure}{1.0\linewidth}

  \includegraphics[width=\linewidth]{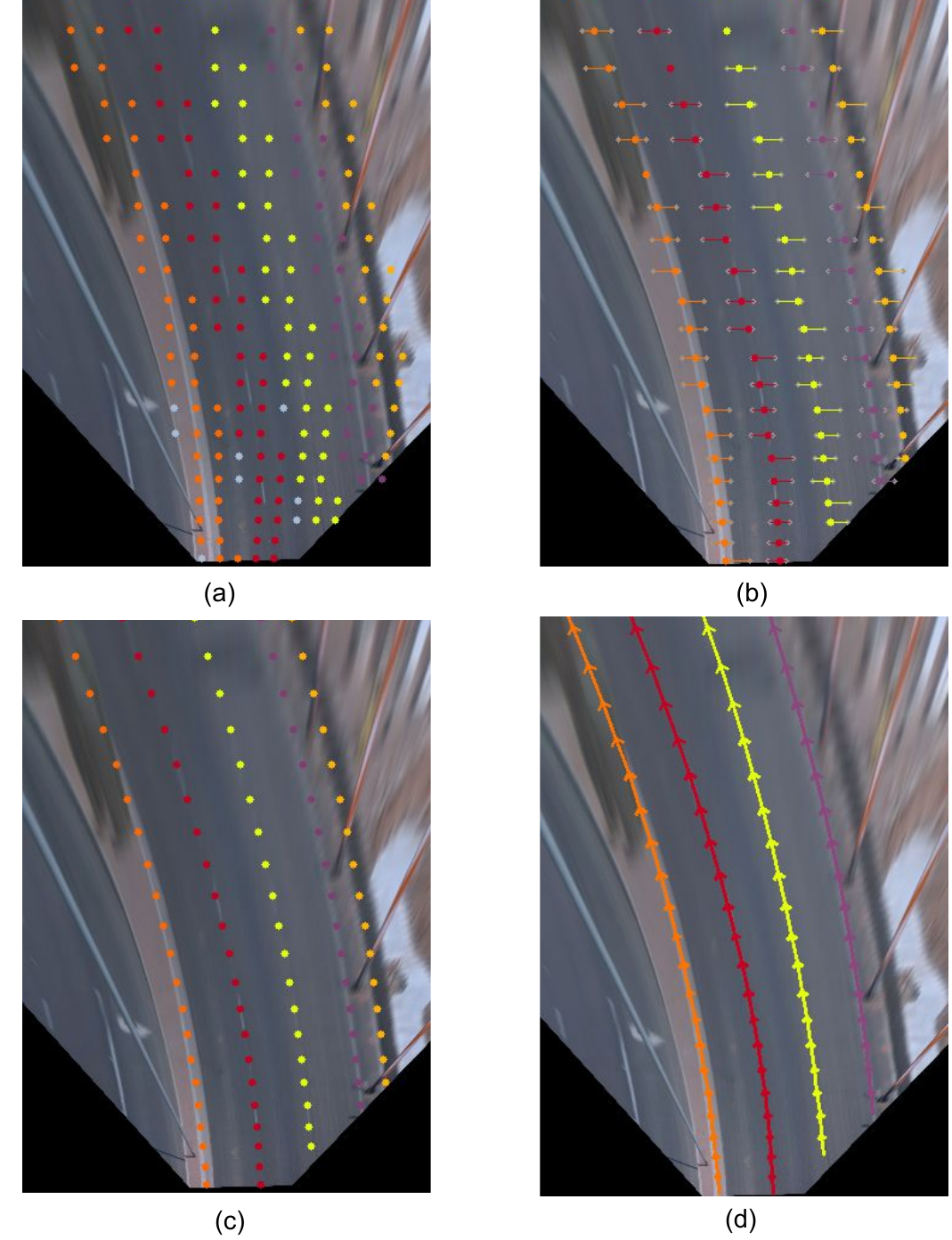}
          \phantomsubcaption

    \label{fig:stages_d}
\end{subfigure}
    \vspace*{-0.6cm}

    \caption{Keypoints over BEV space. Multiple keypoint proposals $K_P$ for a target point at lane (a), regressed offsets $\Delta{x}$ for keypoint proposals (b), strongest proposals $K_S$ after PointNMS (c), predicted directed connections $C$ between strongest keypoints (d). Colors are for visualization and do not indicate class labels.}
    \vspace*{-0.5cm}

\end{figure}

The model utilizes anchor points \( \mathbf{a}_{i,j} \in \mathbf{K}_A^{H_b\times W_b \times 2} \), where \( \mathbf{a}_{i,j} = (x_{i,j}, y_{i,j}) \) for \( i=1,2,\dots,H_b \) and \( j=1,2,\dots,W_b \). Each anchor point \( \mathbf{a}_{i,j} \) corresponds to \( \mathbf{f}_{i,j} = \mathbf{F}_{BEV}[i,j] \), as shown in Fig. \ref{fig:overall}. In the early steps, our model predicts \( \mathbf{M}_{seg} \in \mathbb{R}^{H_b\times W_b\times 1} \) segmentation map to separate foreground/background. The proposal detection module selects anchors as proposal \( \mathbf{K}_P \) that correspond to the highest \( N \) scores in \( \mathbf{M}_{seg} \). 

Each keypoint is essential for lane prediction in bottom-up approaches, and a missing keypoint within a lane may result in the division of a lane into multiple parts. To address this issue, the proposal detection module in GLane3D selects multiple proposals from anchor points for each target keypoint along the lane, as illustrated in Fig. \ref{fig:stages_a}. The proposals are positioned within a lateral distance of up to \( d_x \) from the target lanes. By estimating multiple proposals, the likelihood of missing a keypoint is significantly reduced. The proposal keypoints are then aligned to the target lane by applying the predicted lateral offset, as in Fig. \ref{fig:stages_b}. 

The predicted keypoints are independent, and the lack of relationships between them is addressed by estimating directed connections between consecutive keypoints. During post-processing, we trace these connections to extract lanes. To reduce computational overhead in the connection estimation process, the PointNMS function is employed to retain only the \( S \) most confident keypoints from a set of
\( N \) overlapping candidates, discarding those with lower confidence scores. The relationship head operates on these \( S \) keypoints and outputs the adjacency matrix \( \mathbf{A} \).

\subsection{PV to BEV Space with Special Geometry} \label{sec:special_geometry}

\begin{figure}
  \centering
  \begin{subfigure}{0.75\linewidth}
  \includegraphics[height=3.2cm]{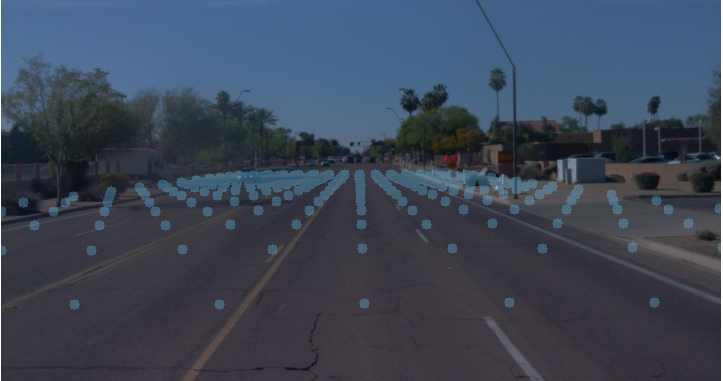}
    \caption{}
    \label{fig:pv_a}
  \end{subfigure}
  \hfill
  \begin{subfigure}{0.24\linewidth}
  \includegraphics[height=3.2cm]{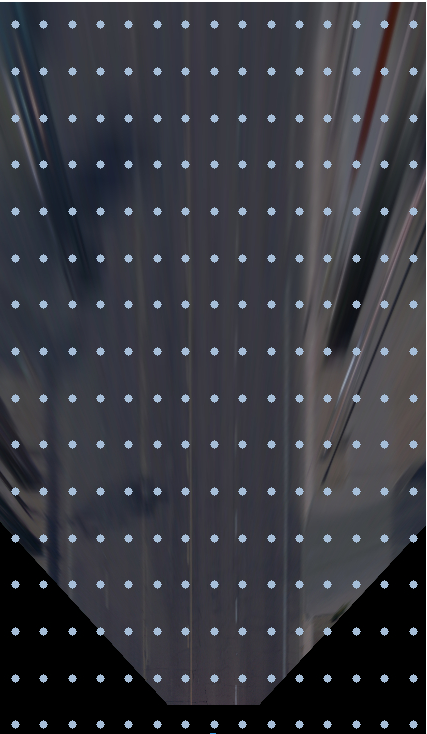}
    \caption{}
    \label{fig:bev_a}
  \end{subfigure}
    \hfill
  \begin{subfigure}{0.75\linewidth}
  \includegraphics[height=3.2cm]{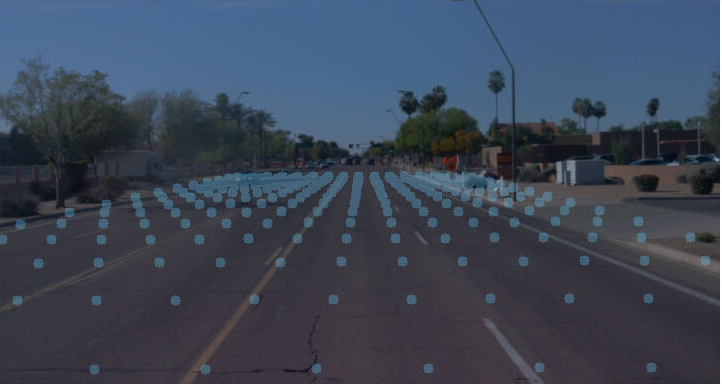}
    \caption{}
    \label{fig:pv_b}
  \end{subfigure}
    \hfill
  \begin{subfigure}{0.24\linewidth}
  \includegraphics[height=3.2cm]{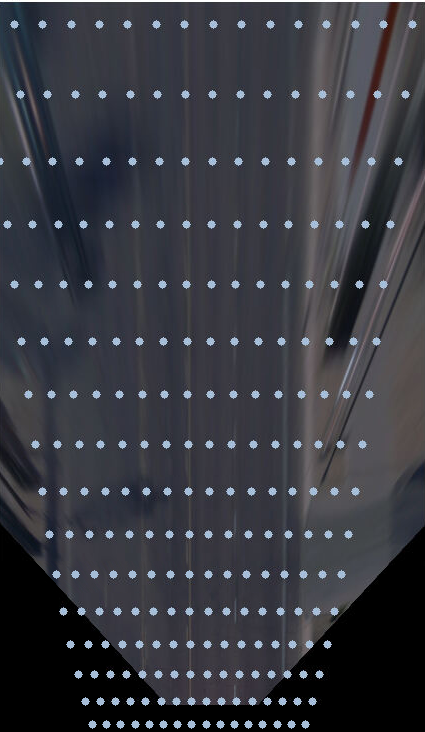}
    \caption{}
    \label{fig:bev_b}
  \end{subfigure}
  \caption{ Uniformly distributed points projected to the front view (a) and represented in the bird's-eye view (b). Customized points projected to the front view (c) and represented in the bird's-eye view (d).}
  \label{fig:short}
    \vspace*{-0.5cm}

\end{figure}

Inverse Perspective Mapping (IPM) projects frontal view (FV) features \( \mathbf{F}_{PV} \) onto the BEV by sampling features from corresponding BEV locations in the FV. These BEV locations are projected to the FV using projection matrices, with the corresponding locations in the FV being determined from the BEV positions. The output of this projection is \( \mathbf{F}_{BEV} \), the feature map in the BEV space. GLane3D utilizes the locations of anchor points \( \mathbf{K}_A^{H_b\times W_b \times 2} \) positioned in the BEV to compute the corresponding PV locations for IPM.

When anchor points are uniformly distributed across the BEV space with equal lateral and longitudinal intervals, the projection onto the FV becomes sparser in regions closer to the ego vehicle and denser in the distant areas, as shown in Fig. \ref{fig:pv_a}. To achieve a denser point distribution in the projected near FV space, we employ a customized approach for distributing \( \mathbf{K}_A^{H_b \times W_b \times 2} \) anchor points across the BEV. In this approach, the longitudinal and lateral distances between anchor points decrease as they approach the ego vehicle, while the number of points in each row remains constant. This strategy results in a denser projection in the FV, as depicted in Fig. \ref{fig:pv_b}. Further details regarding our custom BEV geometry are provided in the supplementary material.

\subsection{Multiple Keypoint Estimation}

GLane3D predicts multiple proposals with the proposal detection module for each target keypoint along the lanes, as in Fig. \ref{fig:stages_a}. The corresponding feature vectors associated with the selected anchor points are collected into \( \mathbf{F}_Q \) as in Eq. \ref{formula:f_q}, which represents the feature vectors corresponding to the proposal points \( \mathbf{K}_P \).
\begin{equation}
    \mathbf{F}_Q = \mathbf{F}_{BEV}[x_i, y_i] \quad \mathbf{k}_i \in \mathbf{K}_P^{N \times 2}.
      \label{formula:f_q}
\end{equation}

The query features \( \mathbf{F}_Q \) are forwarded to a transformer block, which comprises consecutive self-attention and cross-attention layers. In the cross-attention layers, \( \mathbf{F}_{BEV} \), the output of the FV to BEV projection, serves as the memory input. To improve efficiency, we employ deformable attention mechanisms within the cross-attention layers. Four multi-layer perceptrons (MLPs) are utilized to predict classification scores \( \mathbf{S}_{CLS} \), lateral offset \( \Delta x \), height \( z \), and connection feature vector \( \mathbf{f}_c \). Predictions are generated for each proposal point \( \mathbf{k}_i \in \mathbf{K}_P \). To stabilize the training process, we incorporate auxiliary heads that operate on the features from the middle transformer layers.

Following the application of the lateral offset \( \Delta{x} \) to the proposal points \( \mathbf{K}_P \), the proposals are aligned to target keypoints as illustrated in Fig. \ref{fig:stages_b}. To eliminate redundancy while maintaining a high recall rate in keypoint estimation, we employ the PointNMS (Non-Maximum Suppression) operation. This operation retains only the highest-confidence keypoint from among proposals located within a distance \( d_x \) of one another. Proposal confidences are determined by taking the maximum classification score \( \max(s_i) \), where \( s_i \in \mathbf{S}_{CLS} \). As a result, the output keypoints \( \mathbf{K}_S \) from the PointNMS operation exclude any keypoints within a distance \( d_x \) of each other, as in Fig. \ref{fig:stages_c}.

\subsection{Connecting Keypoints}

We extract directed connections \( \mathbf{C} \) between the strongest keypoints \( \mathbf{K}_S \) from the predicted adjacency matrix \( \mathbf{A} \), as in Eq. \ref{eq:con_from_adj}. The directed connections \( \mathbf{C} \) enable a simple lane extraction process. At the detection of the first keypoint with a simple operation, the details are explained in Sec. \ref{sec:postprocessing}.

    \vspace*{-0.3cm}

\begin{equation}
\mathbf{pe}_i = \operatorname{PE}(x_i + \Delta{x}, y_i).
\label{formula:pe}
\end{equation}
\begin{equation}
\mathbf{f}_c^{\prime} = \operatorname{concat}(\mathbf{pe}_i, \mathbf{f}_c).
\end{equation}

The positions of the keypoints are helpful for connection estimation. For this reason, we concatenate the encoded position vector with the feature vector \( \mathbf{f}_c \) as in Eq. \ref{formula:pe}. 

To predict directed connections from origin keypoints to destination keypoints, as illustrated in Fig. \ref{fig:stages_d}, GLane3D processes the feature vectors \( \mathbf{F}_C = (\mathbf{f}_{c_0}^{\prime}, \dots, \mathbf{f}_{c_S}^{\prime}) \) through two separate MLPs. These MLPs output origin keypoint features \( \mathbf{F}_{orig} \in \mathbb{R}^{S \times d} \) and destination keypoint features \( \mathbf{F}_{dest} \in \mathbb{R}^{S \times d} \). Each MLP consists of two linear layers. We then reshape \( \mathbf{F}_{orig} \) to \( \mathbf{F}_{orig}^{\prime} \in \mathbb{R}^{S \times 1 \times d} \) and \( \mathbf{F}_{dest} \) to \( \mathbf{F}_{dest}^{\prime} \in \mathbb{R}^{1 \times S \times d} \). The element-wise multiplication \( \odot \) of these reshaped features produces a tensor of shape \( \mathbb{R}^{S \times S \times d} \). To obtain the adjacency matrix \( \mathbf{A} \in \mathbb{R}^{S \times S} \), a linear layer followed by a sigmoid activation is applied to the element-wise product of \( \mathbf{F}_{orig}^{\prime} \) and \( \mathbf{F}_{dest}^{\prime} \), as formulated in Eq. \ref{formula:conn}.
\vspace{-0.05cm}
\begin{equation}
    \mathbf{C} = \sigma(\operatorname{FC}(\mathbf{F}_{src}^{\prime} \odot \mathbf{F}_{tgt}^{\prime} )).
    \label{formula:conn}
\end{equation}

\subsection{Keypoint Matcher and Loss} \label{sec:keypoint_matcher}

We apply the Hungarian algorithm to match the proposal keypoints with the ground truth keypoints. Since not all \( \mathbf{K}_P \) proposal keypoints are involved in the connection estimation, we run the Hungarian algorithm twice. The first pass is applied to \( \mathbf{K}_P \), while the second pass is applied to the strongest \( \mathbf{K}_S \) keypoints, which are selected through the PointNMS operation.

The Hungarian algorithm considers both the distance between predicted keypoints and ground truth keypoints on the target lane, as well as the similarity of classification probabilities in the cost matrix calculation. The proposal keypoints \( \mathbf{K}_P \) are derived from the anchor points \( \mathbf{K}_A \) and since the starting positions of the points are fixed, we restrict the matching process to prevent the pairing of predicted keypoints with ground truth keypoints that are located far apart. Additionally, the matcher imposes the constraint that keypoints with different longitudinal \( y \)-coordinates cannot be matched.

When applying the Hungarian algorithm to match multiple proposal keypoints \( \mathbf{K}_P \), the ground truth keypoints are repeated \( n \) times to ensure successful matching for all proposal keypoints. However, when matching the strongest keypoints \( \mathbf{K}_S \), we do not repeat the ground truth keypoints, as the PointNMS operation removes any duplicated keypoints.
\vspace{-0.1cm}
\begin{equation}
L_{\text{total}} = w_{\text{kp}} L_{\text{kp}} + w_{\text{r}} L_{\text{r}} + 
w_{\text{cn}} L_{\text{cn}} + w_{\text{c}} L_{\text{c}}.
\label{formula:total_loss}
\end{equation}

We employ Focal Loss \cite{ross2017focal} \( L_{\text{cn}} \) to supervise the connection head. The targets for the connection head are generated based on the matching results. During the early stages of training, keypoints may be missed. In such cases, the next available keypoint in the lane is selected as a positive connection. The regression heads, which estimate the lateral offset \( \Delta_{x} \) and height \( z \), are supervised using L1 loss \( L_{\text{r}} \). The keypoint proposal estimation head is supervised with binary cross-entropy loss \( L_{\text{kp}} \). The classification of keypoints is supervised using cross-entropy loss \( L_{\text{c}} \). GLane3D learns the weights of these losses, denoted as \( w_* \), as proposed in \cite{hu2021fiery}. The total loss function is defined as Eq. \ref{formula:total_loss}.

\subsection{Lane Extraction From Graph} \label{sec:postprocessing}

We extract lanes from the predicted strongest keypoints \( \mathbf{K}_S = (\mathbf{k}_0, \dots, \mathbf{k}_S) \), where each keypoint \( \mathbf{k}_i = (x_i, \Delta{x}, y_i, z) \) represents a keypoint with lateral offset \( \Delta{x} \) and height \( z \), along with the connections between keypoints. The lane extraction algorithm first identifies the start and end keypoints. Start keypoints have no incoming connections but have outgoing connections. Keypoints that satisfy the condition in Eq. \ref{formula:start_point} are classified as start keypoints. Similarly, end keypoints have no outgoing connections but have incoming connections, and we identify them by checking the condition in Eq. \ref{formula:end_point}. The lane extraction process involves finding paths between the start and end keypoints. We use Dijkstra's shortest path algorithm, incorporating the probability of connection \( 1 - \mathbf{A} \) to guide the lane extraction. Since PointNMS has been applied to eliminate duplicate keypoints, the complexity of lane extraction is reduced. Finally, the lane class and lane probability are determined by evaluating the classification probabilities of the keypoints.

\begin{equation}
\sum_{j=1}^{S} \mathbf{C}_{ji} = 0 \quad \text{and} \quad \sum_{j=1}^{S} \mathbf{C}_{ij} > 0.
\label{formula:start_point}
\vspace{-0.3cm}
\end{equation}

\begin{equation}
    \sum_{j=1}^{S} \mathbf{C}_{ji} > 0 \quad \text{and} \quad \sum_{j=1}^{S} \mathbf{C}_{ij} = 0.
    \label{formula:end_point}
\end{equation}

\section{Experiments}
\label{sec:experiments}
\begin{table*} 
    \begin{center}
         \begin{tabular}{ c | l |  l  |  c |  c c | c c } 
         \Xhline{1.3pt}
         Dist. & Methods & Backbone & F1-Score$\uparrow$  &\begin{tabular}{@{}c@{}}X-error \\  near(m) $\downarrow$  \end{tabular}  & \begin{tabular}{@{}c@{}}X-error \\  far(m) $\downarrow$  \end{tabular}   & \begin{tabular}{@{}c@{}}Z-error \\  near(m) $\downarrow$ \end{tabular} & \begin{tabular}{@{}c@{}}Z-error \\  far(m) $\downarrow$  \end{tabular}  \\ [0.5ex] 
         \hhline{========}
        
         \parbox[t]{2mm}{\multirow{10}{*}{\rotatebox[origin=c]{90}{\textbf{\textit{1.5m}}}}}
         & PersFormer \cite{chen2022persformer} &  EffNet-B7 & 50.5 & 0.485 & 0.553 & 0.364  & 0.431  \\
         & BEV-LaneDet \cite{wang2023bev}& ResNet-34  & 58.4 & 0.309 & 0.659 & 0.244  & 0.631 \\
         & Anchor3DLane \cite{huang2023anchor3dlane}& EffNet-B3 & 53.7 & 0.276 & 0.311 & 0.107  & 0.138 \\
         & LATR \cite{luo2023latr}& ResNet-50 & 61.9 & 0.219 & 0.259 & 0.075  & 0.104 \\
         & LaneCPP \cite{pittner2024lanecpp}& EffNet-B7 & 60.3 & 0.264 & 0.310 & 0.077 & 0.117 \\
         & PVALane \cite{zheng2024pvalane}& ResNet-50 & 62.7 & 0.232 & 0.259 & 0.092  & 0.118 \\
         & PVALane \cite{zheng2024pvalane}& Swin-B & 63.4 & 0.226 & 0.257 & 0.093  & 0.119 \\  
         \cline{2-8}
         & \mycc GLane3D-Lite (Ours) & \mycc ResNet-18 & \mycc 61.5 & \mycc 0.221 & \mycc 0.252 & \mycc 0.073 & \mycc 0.101 \\
          \cline{2-8}
         & \mycc GLane3D-Base (Ours) & \mycc ResNet-50 & \mycc \underline{63.9} & \mycc \underline{0.193} & \mycc \underline{0.234} &\mycc \underline{0.065}  &\mycc \underline{0.090} \\
        \cline{2-8}
        &\mycc GLane3D-Large (Ours) &\mycc Swin-B &\mycc \textbf{66.0} &\mycc \textbf{0.170} &\mycc \textbf{0.203} &\mycc \textbf{0.063}  &\mycc \textbf{0.087} \\

        \Xhline{1.3pt}

         \parbox[t]{2mm}{\multirow{8}{*}{\rotatebox[origin=c]{90}{\textbf{\textit{0.5m}}}}} & PersFormer \cite{chen2022persformer} & EffNet-B7  & 36.5 & 0.343 & 0.263 & 0.161  & 0.115  \\
          & Anchor3DLane \cite{huang2023anchor3dlane} & EffNet-B3  & 34.9 & 0.344 & 0.264 & 0.181  & 0.134   \\
          & PersFormer \cite{chen2022persformer} &  ResNet-50 & 43.2 & 0.229 & 0.245 & 0.078  & 0.106  \\
          & LATR \cite{luo2023latr}& ResNet-50 & 54.0 & 0.171 & 0.201 & 0.072  & 0.099 \\
          & DV-3DLane(Camera) \cite{luo2024dv}& ResNet-34 & 52.9 & 0.173 & 0.212 & 0.069  & 0.098 \\
         \cline{2-8} &\mycc  GLane3D-Lite (Ours) &\mycc ResNet-18 &\mycc 53.8 &\mycc 0.182 &\mycc 0.206 &\mycc 0.070 &\mycc 0.095 \\
          \cline{2-8} &\mycc  GLane3D-Base (Ours) &\mycc ResNet-50 &\mycc \underline{57.9} &\mycc \underline{0.157} &\mycc \underline{0.179} &\mycc \underline{0.067}  &\mycc \underline{0.087} \\
        \cline{2-8} &\mycc  GLane3D-Large (Ours) &\mycc Swin-B &\mycc \textbf{61.1} &\mycc \textbf{0.142} &\mycc \textbf{0.167} &\mycc \textbf{0.061}  &\mycc \textbf{0.084} \\
        \Xhline{1.3pt}

        \end{tabular}
       \caption{Quantitative results on OpenLane\cite{chen2022persformer} Dataset}
       \label{tab:openlane}
    \end{center}
      \vspace*{-\baselineskip}

\end{table*}

We evaluate our models on two different datasets, real-world dataset OpenLane, and the synthetic dataset Apollo. Both datasets contain ground truth lanes in 3D space and camera parameters

\subsection{Datasets}
\textbf{OpenLane}\cite{chen2022persformer} is a large-scale dataset constructed from the Waymo Open Dataset \cite{sun2020scalability}, comprises 200k frames, with 150k in the training set and 50k in the test set, drawn from a total of 1,000 scenes. This dataset includes 880,000 lane annotations, all provided in 3D space. The test set scenes are divided into eight distinct categories: Up \& Down, Curve, Extreme Weather, Night, Intersection, and Merge \& Split. A subset containing 300 scenes, referred to as Lane300, is utilized specifically for ablation studies.

\textbf{Apollo}\cite{guo2020gen} is a small synthetic dataset, that contains 10,500 frames captured from highway, urban, and rural environments. It is organized into three subsets: standard, rare scenes, and visual variations.

\subsection{Metric}
We adopt the evaluation metrics proposed by Gen-LaneNet \cite{guo2020gen} for both 3D datasets. These metrics assess the Euclidean distance at uniformly spaced points along the y-axis within a range of 0–100 meters. A successful match between a predicted lane and the ground truth lane is defined by at least 75\% of points in the longitudinal (y) direction falling within a specified distance threshold. The default distance threshold is set at 1.5 meters; however, due to the relative largeness of this threshold, we also include an additional 0.5-meter threshold for evaluation.

\subsection{Implementation Details}

\begin{table}
    \centering
    \begin{tabular}{c| c | c|c|c}
        \Xhline{1.3pt}
         \begin{tabular}{@{}c@{}}Point \\  NMS \end{tabular} & \begin{tabular}{@{}c@{}}Multiple \\  Proposal \end{tabular} & \begin{tabular}{@{}c@{}}Custom \\  BEV \end{tabular} & F1(\%)  & Gain \\
         \hhline{=====}

         & & &  66.6 &  \\
         \checkmark & & &  69.2 & +2.6 \\
         & \checkmark  & & 42.7 & - \\
         \checkmark & \checkmark  & &  71.6 & +5.0 \\
         \checkmark & \checkmark  & \checkmark  &  72.0 & +5.4 \\

         \Xhline{1.3pt}

    \end{tabular}
    \caption{Performance gain for parts of GLane3D on OpenLane300 using PointNMS, Multiple Keypoint Proposal, Cutom BEV Geometry}
    \label{tab:ablation}
      \vspace*{-\baselineskip}
\end{table}

We use an input size of $384 \times 720$ for lite and base models, and $512 \times 960$ for the large model. The projected BEV dimensions are $56 \times 32$, $56 \times 64$, and $72 \times 128$, respectively. The maximum number of proposal keypoints ($S$) is 256 for lite and base models, with repeats ($n$) set to 2, and 384 and 4 for the large model. We employ the Adam optimizer with a learning rate of $3 \times 10^{-4}$, warm-up, and cosine annealing. Training runs for 24 epochs on OpenLane version 1.2,  which was also used in previous works, and 300 epochs on Apollo, with a batch size of 16.

\subsection{Ablation Studies}

Tab.\ref{tab:ablation} shows that, while the PointNMS alone can increase the score, multiple keypoint proposals alone do not increase the the F1 score but decrease the F1 score. But when we used PointNMS and multiple proposal keypoint approach together, detection performance is increased. The performance degredation when only multiple keypoint proposal introduced is result of ambiguity in connection graph. Since GLane3D pushes that connect keypoint with only one predecessor keypoint and only one successor keypoint, multiple keypoint at same location causes ambiguity. PointNMS solves multiple keypoint at same location ambiguity while reducing computation overhead.
Performance gain comes with custom BEV geometry at IPM proves contribution of increasing density in ego near locations in BEV.

Tab.\ref{tab:ablation2} shows that the contribution of increasing the number of keypoints $S$ 
saturates after 256 nodes. Beyond this optimal point, the performance gain diminishes, while the computational overhead of additional proposal keypoints reduces the FPS. This observation demonstrates the impact of foreground/background separation with the proposal estimation head on execution speed.

\begin{table*}
    \begin{center}
         \begin{tabular}{ l |  l  |  c |  c | c | c | c | c } 
        \Xhline{1.3pt}
         Methods & Backbone & \begin{tabular}{@{}c@{}}Up \& \\ Down $\uparrow$  \end{tabular}   & Curve $\uparrow$ & \begin{tabular}{@{}c@{}}Extreme \\  Weather $\uparrow$  \end{tabular}   & Night $\uparrow$  & Intersection $\uparrow$  &  \begin{tabular}{@{}c@{}}Merge \\  Split $\uparrow$  \end{tabular} \\ [0.5ex] 
        \hhline{========}
        
         PersFormer \cite{chen2022persformer} &  EffNet-B7 & 42.4 & 55.6 & 48.6 & 46.6  & 40.0 & 50.7 \\
         BEV-LaneDet \cite{wang2023bev}& ResNet-34  & 48.7 & 63.1 & 53.4 & 53.4  & 50.3 & 53.7 \\
         PersFormer \cite{chen2022persformer} &  ResNet-50 & 46.4 & 57.9 & 52.9 & 47.2  & 41.6 & 51.4 \\

         LATR \cite{luo2023latr}& ResNet-50 & 55.2 & 68.2 & 57.1 & 55.4  & 52.3 & 61.5 \\
         LaneCPP \cite{pittner2024lanecpp}& EffNet-B7 & 53.6 & 64.4 & 56.7 & 54.9  & 52.0 & 58.7 \\
         PVALane \cite{zheng2024pvalane}& ResNet-50 & 54.1 & 67.3 & 62.0 & 57.2  & 53.4 & 60.0 \\
         \cline{1-8} \rowcolor{lightgray} Glane3D-Lite (Ours) & ResNet-18 & 55.6 & 69.1 & 56.6 & 56.6 & 52.9 & 61.3\\
          \cline{1-8} \rowcolor{lightgray} GLane3D-Base (Ours) & ResNet-50 & \underline{58.2} & \underline{71.1} & \underline{60.1} & \underline{60.2}  & \underline{55.0} & \underline{64.8} \\
        \cline{1-8} \rowcolor{lightgray} GLane3D-Large (Ours) & Swin-B & \textbf{61.7} & \textbf{72.7} & \textbf{63.8} & \textbf{62.0}  & \textbf{57.9} & \textbf{67.7} \\
        \Xhline{1.3pt}
        \end{tabular}
       \caption{Quantitative results per category on OpenLane\cite{chen2022persformer} Dataset}
       \label{tab:openlane_category}
    \end{center}
    \vspace*{-0.5cm}
\end{table*}

\subsection{Results on OpenLane}

\begin{figure}[b]
        \includegraphics[width=\linewidth]{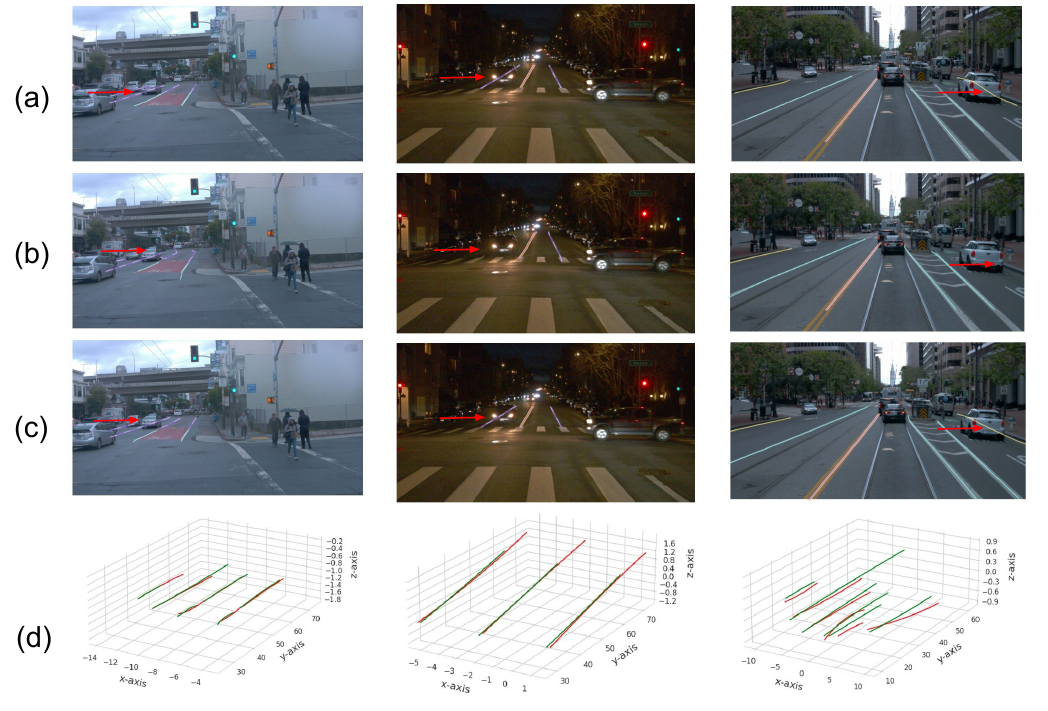}
    \caption{Qualitative evaluation on OpenLane val set. The rows illustrate (a) ground truth 3D lanes, prediction from (b) LATR\cite{luo2023latr} and (c) GLane3D with 2D projection, respectively. Here, different colors indicate specific categories. Row (d) demonstrates the ground truth (red) and prediction of GLane3D (green) in 3D space. Best viewed in color (zoom in for details).}
        \vspace*{-0.2cm}

\end{figure}

The results in Tab.\ref{tab:openlane} demonstrate that our GLane3D model consistently outperforms previous state-of-the-art (SoTA) methods, particularly in F1 score performance. Specifically, we achieved a 1.2\% improvement over a model that uses a ResNet-50 backbone, and a notable 2.6\% increase when using a Swin-B backbone, outperforming PVALane with the same configuration. In terms of localization, GLane3D exhibits superior accuracy in the x dimension and z dimension. Notably, even with a stricter 0.5-meter matching threshold, GLane3D maintains a strong advantage over other SoTA models, demonstrating its robustness under various evaluation conditions.

\begin{table}[t]
    \centering
    \begin{tabular}{c|c|c}
        \Xhline{1.3pt}
         Number of Keypoints & F1-Score(\%) $\uparrow$ & FPS $\uparrow$ \\
         \hhline{===}

         128 &  55.5 & 28.5  \\
         256 &  72.0 & 27.8 \\
         512 & 72.4 & 25.9 \\
         1024 & 72.4 & 21.0 \\

         \Xhline{1.3pt}

    \end{tabular}
    \caption{Effect of number of Keypoints to performance and speed}
    \label{tab:ablation2}
      \vspace*{-0.4cm}

\end{table}

In a category-based analysis at Tab.\ref{tab:openlane_category}, GLane3D shows improvements across all categories when compared to previous SoTA methods. Results from the Up \& Down category highlight the effectiveness of the IPM-based projection in learning non-flat surfaces, which enhances the model’s ability to handle various terrains. Additionally, the flexibility of the keypoint-based representation allows GLane3D to predict curved lanes with a 3.8\% improvement, while also yielding 4.8\% and 1.6\% F1 score gains in handling merge-split and intersection cases, respectively. These results underscore the benefits of the keypoint-based approach, particularly for complex lane shapes and intersections.

GLane3D also delivers an effective balance of accuracy and efficiency. The GLane3D-Lite model achieves a remarkable 62.2 FPS, surpassing most models in terms of F1 score and making it highly practical for real-time applications as in Tab.\ref{tab:f1_vs_fps}. Meanwhile, GLane3D-Base outperforms LATR-Lite, achieving a 2.4\% higher F1 score at a comparable frame rate. This efficiency reduces the computational demands for in-vehicle deployment, ultimately lowering costs and increasing accessibility for automotive customers.

While lidar sensor prices have decreased in recent years, they remain significantly higher than camera costs. As a result, GLane3D primarily focuses on camera-based processing rather than complex camera+lidar fusion. Nonetheless, we trained and evaluated GLane3D with combined camera and lidar inputs to showcase its adaptability. For lidar processing, we utilized SECOND \cite{yan2018second} as the lidar feature extractor, combining the projected IPM feature with the lidar feature in subsequent layers. Despite not implementing advanced fusion techniques, GLane3D still slightly outperforms previous models in F1 score as in Tab.\ref{tab:camera_lidar_fusion}, underscoring its versatility and robustness across input configurations.

\begin{table}
    \centering
    \begin{tabular}{c|c|c|c}
        \Xhline{1.3pt}
         Methods & Backbone & F1-Score(\%) $\uparrow$ & FPS  $\uparrow$ \\
         \hhline{====}

         Persformer & EffNet-B7 & 50.5 & 20.3 \\
         LATR-Lite & ResNet-50 & 61.5 & 23.5 \\
         LATR &  ResNet-50 & 61.9 & 15.2 \\
       GLane3D-Lite & ResNet-18 & 61.5 & \textbf{62.2} \\
       GLane3D-Base & ResNet-50 & \underline{63.9} & \underline{27.8}\\
       GLane3D-Large & Swin-B & \textbf{66.0} & 22.1\\

         \Xhline{1.3pt}

    \end{tabular}
    \caption{Model performance (F1) vs Execution Speed (FPS)}
    \label{tab:f1_vs_fps}
      \vspace*{-0.4cm}

\end{table}

\subsection{Results on Apollo}

Tab.\ref{tab:apollo} shows quantitative results at Apollo dataset. GLane3D outperforms previous methods in balanced scenes and rare scenes \textit{w.r.t.} F1 score. Though localization errors are close to saturation, our method achieves lower localization error for most localization errors. More detailed results are in supplementary material.

\begin{table*}
     \begin{tabular}{ l | c |  c  c |  c  c || c | c c | c c } 
    \Xhline{1.3pt}
      \multirow{2}{*}{\parbox{1cm}{Method }} & \multicolumn{5}{c ||}{ \textbf{Balanced Scenes}} & \multicolumn{5}{c }{ \textbf{Rare Scenes}} \\
      \cline{2-11}& F1(\%) &\begin{tabular}{@{}c@{}}X-err\\near(m)  \end{tabular}  & \begin{tabular}{@{}c@{}}X-err \\  far(m)   \end{tabular}   & \begin{tabular}{@{}c@{}}Z-err \\  near(m) \end{tabular} & \begin{tabular}{@{}c@{}}Z-err \\  far(m)  \end{tabular}  & F1(\%) &\begin{tabular}{@{}c@{}}X-err \\  near(m)  \end{tabular}  & \begin{tabular}{@{}c@{}}X-err \\  far(m)  \end{tabular}   & \begin{tabular}{@{}c@{}}Z-err \\  near(m)  \end{tabular} & \begin{tabular}{@{}c@{}}Z-err \\  far(m)   \end{tabular}  \\ [0.5ex] 
    \hhline{===========}

     PersFormer\cite{chen2022persformer} & 92.9  & 0.054 & 0.356 & 0.010  & 0.234 & 87.5   & 0.107 & 0.782 & 0.024  & 0.602  \\
      BEVLaneDet \cite{wang2023bev} & 96.9 & \textbf{0.016} & \textbf{0.242} & 0.020  & 0.216  & \underline{97.6} & \textbf{0.031} & \textbf{0.594} & 0.040  & 0.556   \\
     LaneCPP \cite{pittner2024lanecpp} &  \underline{97.4} & 0.030 & 0.277 & \underline{0.011}  & \textbf{0.206} & 96.2 & 0.073 & 0.651 & \underline{0.023}  & \underline{0.543} \\
     LATR \cite{luo2023latr}&  96.8 & 0.022 & 0.253 & \textbf{0.007}  & \textbf{0.202} &  96.1 &  0.050 & \underline{0.600} & \textbf{0.015}  & \textbf{0.532} \\
     DV-3DLane \cite{luo2024dv}&  96.4  & 0.046 & 0.299 & 0.016  & \underline{0.213} &  95.5 & 0.071 & 0.664 & 0.025  & 0.568  \\

    \rowcolor{lightgray} GLane3D (Ours) &  \textbf{98.1} & \underline{0.021} & \underline{0.250} & \textbf{0.007}  & \underline{0.213} &  \textbf{98.4}  & \underline{0.044} & 0.621 & 0.023  & 0.566 \\
    \Xhline{1.3pt}

    \end{tabular}
   \caption{Quantitative results on Apollo\cite{guo2020gen} 3D Synthetic Dataset}
   \label{tab:apollo}
         \vspace*{-0.2cm}
\end{table*}

\subsection{Cross Dataset Evaluation}

\begin{table}
     \begin{tabular}{l | l | l | c |  c  } 
    \Xhline{1.3pt}
     $D_t$ & Methods & F1 & X-near/far & Z-near/far\\ [0.5ex] 
    \hhline{=====}
    
     \parbox[t]{2mm}{\multirow{3}{*}{\rotatebox[origin=c]{90}{\textbf{\textit{1.5m}}}}} & PersFormer\cite{chen2022persformer} &  \underline{53.2} & 0.407/0.813 & \underline{0.122}/\textbf{0.453}\\
     & LATR \cite{luo2023latr}&  34.3 & \underline{0.327}/\underline{0.737} & 0.142/0.500 \\
     
    & GLane3D & \textbf{58.9} & \textbf{0.289}/\textbf{0.701} & \textbf{0.086}/\underline{0.479} \\

    \Xhline{1.3pt}

     \parbox[t]{2mm}{\multirow{3}{*}{\rotatebox[origin=c]{90}{\textbf{\textit{0.5m}}}}} & PersFormer\cite{chen2022persformer} &  \underline{17.4} & 0.246/0.381 & \underline{0.098}/\underline{0.214}\\
     & LATR \cite{luo2023latr}&  19.0 & \underline{0.201}/\underline{0.313} & 0.116/0.220 \\
    & GLane3D & \textbf{42.6} & \textbf{0.162}/\textbf{0.296} & \textbf{0.063}/\textbf{0.198} \\
        \Xhline{1.3pt}

        \end{tabular}
   \caption{Cross Dataset Evaluation on Apollo\cite{guo2020gen} Balanced Scenes}
   \label{tab:apollo_cross}
         \vspace*{-0.2cm}

\end{table}

\begin{figure}[b]
        \includegraphics[width=\linewidth]{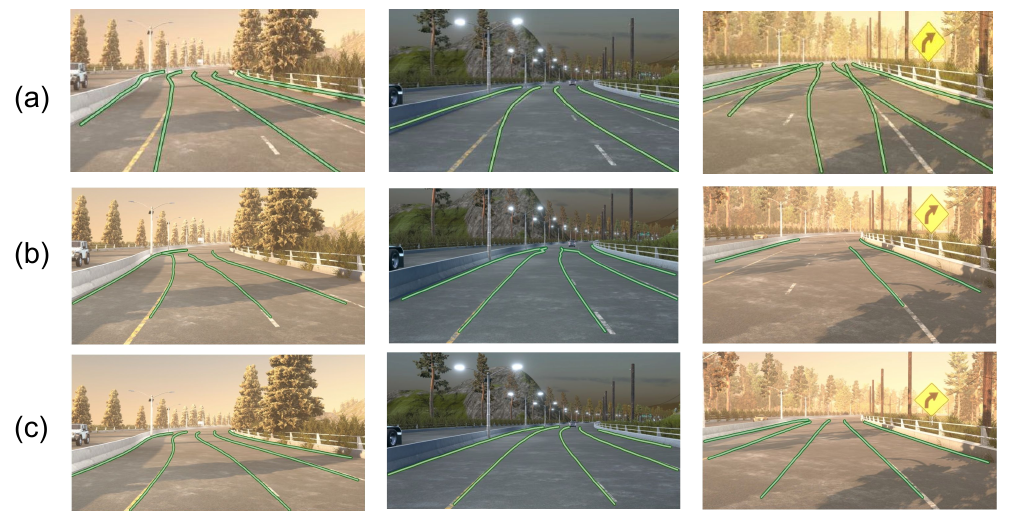}
    \caption{Qualitative results on cross dataset evaluation on Apollo validation set of Balanced Scenes. The rows illustrate prediction from (a) PersFormer\cite{chen2022persformer}, (b) LATR\cite{luo2023latr} and (c) GLane3D with 2D projection, respectively. Best viewed in color (zoom in for details).}
            \label{fig:cross_eval}
  \vspace*{-\baselineskip}

\end{figure}

To assess the generalization ability of the methods, we evaluated models trained on the OpenLane dataset using the Apollo dataset. This approach allows us to observe inference results with images taken from different camera locations (pitch, roll, yaw, and x, y, z positions), in unseen environments, and containing unseen lane structures. Tab.\ref{tab:apollo_cross} shows substantial improvement in the generalization performance of GLane3D, including F1 score and localization errors. The smooth lane predictions of GLane3D, compared to the lane predictions of LATR, demonstrate that top-down approaches are not as effective at generalization, as seen in Fig.\ref{fig:cross_eval}. Another reason for the performance gap between LATR and GLane3D is the absence of explicit FV to BEV projection. Qualitative results reveal a lack of localization in PersFormer during cross-dataset inference, especially in near areas. Since the distance threshold in F1 score calculation is 1.5m, the detection performance results in Tab.\ref{tab:apollo_cross} are not adversely affected.

\begin{table}
     \begin{tabular}{ l |  l | c  | c | c } 
    \Xhline{1.3pt}
     $D_t$ & Methods&F1& X-near/far & Z-near/far  \\ [0.5ex] 
    \hhline{=====}
    
     \parbox[t]{2mm}{\multirow{3}{*}{\rotatebox[origin=c]{90}{\textbf{\textit{1.5m}}}}} & $\operatorname{M}^2$-3DLaneNet &55.5& 0.283/0.256 &0.078/0.106  \\
     & DV-3DLane \cite{luo2024dv}&\textbf{66.8}& \underline{0.115}/\underline{0.134}& \textbf{0.024}/\textbf{0.049} \\

    & GLane3D* &63.9&0.193/0.234&0.065/\underline{0.090} \\
    & GLane3D\ddag&\underline{66.6}&\textbf{0.095}/\textbf{0.112}& \underline{0.026}/\textbf{0.049} \\

    \Xhline{1.3pt}

     \parbox[t]{2mm}{\multirow{3}{*}{\rotatebox[origin=c]{90}{\textbf{\textit{0.5m}}}}} & $\operatorname{M}^2$-3DLaneNet &48.2& 0.217/0.203 &0.076/0.103  \\
     & DV-3DLane \cite{luo2024dv}&\underline{65.2}& \underline{0.082}/\underline{0.101}& \underline{0.028}/\textbf{0.048} \\

    & GLane3D* &57.9&0.157/0.179&0.067/\underline{0.087} \\
    & GLane3D\ddag&\textbf{65.6}&\textbf{0.068}/\textbf{0.091}& \textbf{0.025}/\textbf{0.048} \\
        \Xhline{1.3pt}

        \end{tabular}
   \caption{Camera + LiDAR Fusion. GLane3D* represents the camera-only model, while GLane3D\ddag indicates the camera and LiDAR fusion model}
   \label{tab:camera_lidar_fusion}
         \vspace*{-0.5cm}

\end{table}

    \vspace*{-0.3cm}
\section{Conclusion}

In this work, we propose GLane3D, a keypoint-based 3D lane detection method that uses directed connection estimation for efficient lane extraction. By enhancing keypoint detection with multiple proposals and PointNMS, we reduce computational overhead. Inverse Perspective Mapping (IPM) with customized BEV locations improves sparsity near the ego vehicle and alleviates saturation in distant areas. GLane3D’s approach improves generalization, as shown by cross-dataset evaluations. It outperforms state-of-the-art methods on both OpenLane and Apollo datasets, achieving high F1 scores in camera+Lidar fusion.

{
    \small
    \bibliographystyle{ieeenat_fullname}
    \bibliography{main}
}

\clearpage
\setcounter{page}{1}
\maketitlesupplementary

  \lstdefinestyle{pythonstyle}{
    language=Python,
    basicstyle=\ttfamily\footnotesize,
    keywordstyle=\color{blue}\bfseries,
    stringstyle=\color{green},
    commentstyle=\color{gray},
    showstringspaces=false,
    frame=single,
    numbers=left,
    numberstyle=\tiny,
    breaklines=true
}

\begin{figure}[ht]
    \centering
    \lstset{style=pythonstyle}
    \begin{lstlisting}
def PointNMS(points_x, points_y, scores,
             thresh_x, thresh_y, r=10):
    x1 = points_x * r - (r / 2) * thresh_x
    x2 = points_x * r + (r / 2) * thresh_x
    y1 = points_y * r - (r / 2) * thresh_y
    y1 = points_y * r + (r / 2) * thresh_y
    boxes = stack(round(x1), round(y1),
                  round(x2), round(y2))

    keep = BoxNMS(boxes, scores, iou_thresh=0.1)
    return keep
\end{lstlisting} 

\caption{Python code of PointNMS function}
\label{fig:point_nms code}
    \vspace*{-0.5cm}

\end{figure}

\section{Hyperparameter Relationship: $N$ and $S$ in Proposal Selection}

The number of proposals, \( N \), serves as a hyperparameter in GLane3D, determining the total proposals selected from the set of anchor keypoints \( \mathbf{K}_A \). Our model leverages multiple proposals per keypoint to effectively represent the target lane. These proposals are refined using PointNMS, a function that retains the strongest \( S \) non-overlapping keypoints. Since each keypoint is chosen from \( n \) proposals, where \( n \) represents the number of proposals per target keypoint, the relationship \( N = S \times n \) is adhered to during the hyperparameter selection process.

\begin{table*}[b]
    \begin{center}
         \begin{tabular}{ c | l |  l  |  c | c |  c c | c c } 
        \Xhline{1.3pt}
         Dist. & Methods & Backbone & F1-Score$\uparrow$  & AP(\%) $\uparrow$ &\begin{tabular}{@{}c@{}}X-error \\  near(m) $\downarrow$  \end{tabular}  & \begin{tabular}{@{}c@{}}X-error \\  far(m) $\downarrow$  \end{tabular}   & \begin{tabular}{@{}c@{}}Z-error \\  near(m) $\downarrow$ \end{tabular} & \begin{tabular}{@{}c@{}}Z-error \\  far(m) $\downarrow$  \end{tabular}  \\ [0.5ex] 
        \hhline{=========}
        
         \parbox[t]{2mm}{\multirow{4}{*}{\rotatebox[origin=c]{90}{\textbf{\textit{1.5m}}}}}
         & PersFormer \cite{chen2022persformer} &  EffNet-B7 & 53.2 & - & 0.407 & 0.813 & 0.122  & \textbf{0.453}  \\
         & LATR \cite{luo2023latr}& ResNet-50 & 34.3 & 51.2 & 0.327 & 0.737 & 0.142  & 0.501 \\
         \cline{2-9} & \mycc GLane3D-Base (Ours) &\mycc ResNet-50 &\mycc \underline{54.9} &\mycc \textbf{64.0} &\mycc \textbf{0.255} &\mycc \textbf{0.694} &\mycc \textbf{0.078} &\mycc \underline{0.464} \\
        \cline{2-9} &\mycc GLane3D-Large (Ours) & \mycc Swin-B &\mycc \textbf{58.9} &\mycc \underline{64.9} &\mycc\underline{0.289} &\mycc \underline{0.701} &\mycc \underline{0.086}  &\mycc 0.479 \\
          
        \Xhline{1.3pt}

         \parbox[t]{2mm}{\multirow{4}{*}{\rotatebox[origin=c]{90}{\textbf{\textit{0.5m}}}}} & PersFormer \cite{chen2022persformer} & EffNet-B7  & 17.4 & - & 0.246 & 0.381 & 0.098  & 0.214  \\
          & LATR \cite{luo2023latr} & ResNet-50  & 19.0 & 27.8 & 0.201 & 0.313 & 0.116  & 0.220   \\
          \cline{2-9} &\mycc GLane3D-Base (Ours) &\mycc ResNet-50 &\mycc \underline{40.7} &\mycc \underline{44.2} &\mycc \textbf{0.135} &\mycc \underline{0.297} &\mycc \textbf{0.055}  &\mycc \textbf{0.194} \\
        \cline{2-9} &\mycc  GLane3D-Large (Ours) &\mycc Swin-B &\mycc \textbf{42.6} &\mycc \textbf{44.4} &\mycc \underline{0.162} &\mycc \textbf{0.296} &\mycc \underline{0.063}  &\mycc \underline{0.198} \\
        \Xhline{1.3pt}

        \end{tabular}
       \caption{Extended Cross-Dataset Evaluation on the Balanced Scenes of the Apollo Dataset \cite{guo2020gen}.}
       \label{tab:extend_cross_dataset}
    \end{center}
      \vspace*{-\baselineskip}

\end{table*}

\section{Cross Dataset Evaluation}

\cref{tab:extend_cross_dataset} provides an extended version of \cref{tab:apollo_cross}, presenting a more comprehensive comparison. The results demonstrate that GLane3D achieves superior performance in cross-dataset evaluations.  

The gap in the 0.5m thresholded F1-score highlights the superior generalization capability of our model. Specifically, while GLane3D achieves a +5.7 improvement in F1-score under the 1.5m threshold, the notable +25.2 improvement in the 0.5m threshold further underscores its enhanced generalization compared to previous methods.  

\begin{figure*}
  \centering
    \begin{subfigure}{0.0\linewidth}
            \phantomsubcaption
        \label{fig:pre_nms}
    \end{subfigure}
        \begin{subfigure}{0.0\linewidth}
        \phantomsubcaption
        \label{fig:nms_box}

    \end{subfigure}
    
  \begin{subfigure}{1.0\linewidth}

  \includegraphics[width=\linewidth]{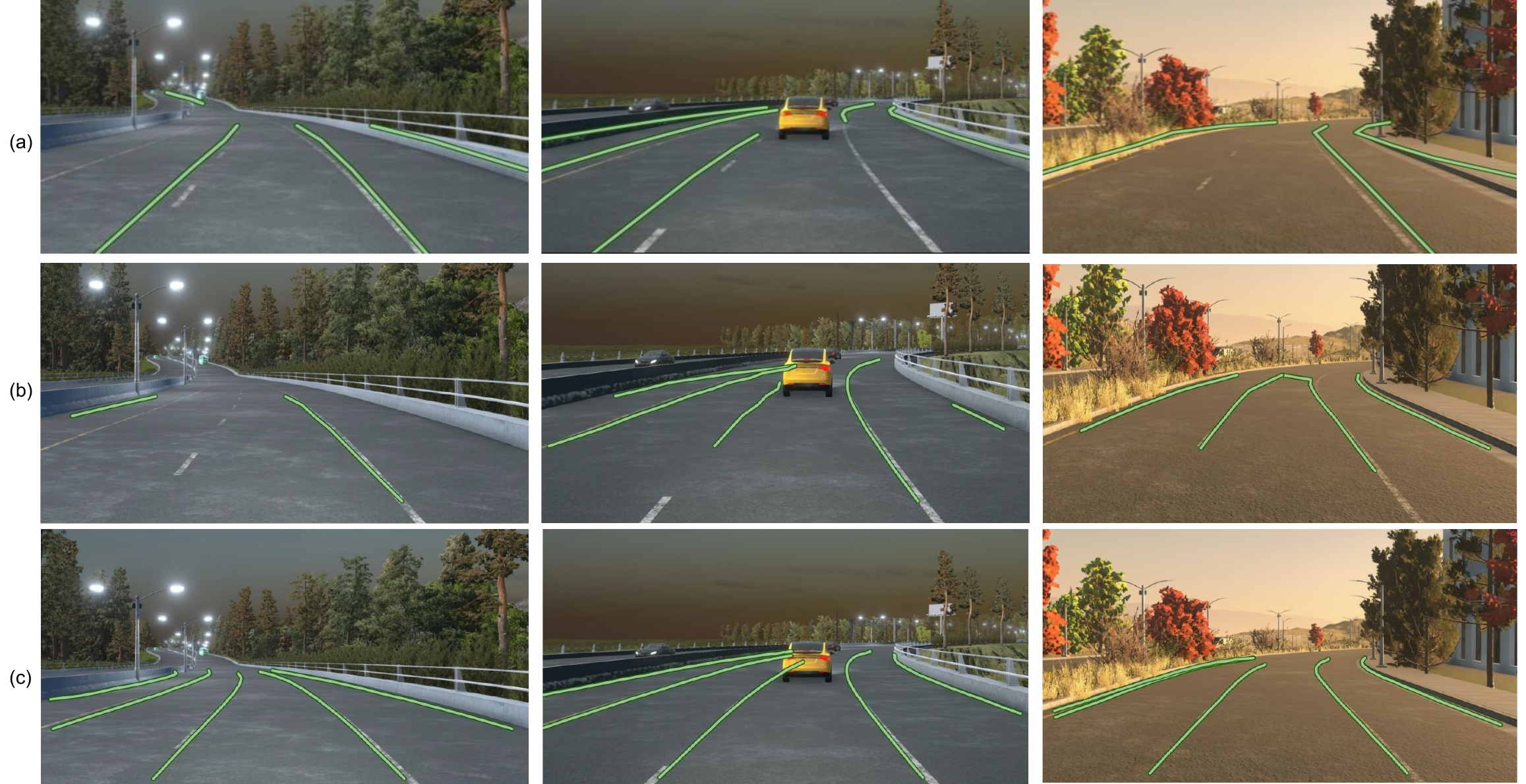}
          \phantomsubcaption

    \label{fig:post_nms}
\end{subfigure}
    \caption{Qualitative results on cross dataset evaluation on Apollo validation set of Balanced Scenes. The rows (a), (b), (c) illustrate prediction from PersFormer\cite{chen2022persformer}, LATR\cite{luo2023latr} and GLane3D with 2D projection, respectively.}
    \label{fig:qual_cross_dataset}

\end{figure*}

The qualitative results in \cref{fig:qual_cross_dataset} help to explain the significant F1-score gap under the 0.5m matching threshold. These results reveal that the localization performance of previous methods is inferior to that of GLane3D, particularly in precise lane representation.

\section{PointNMS Implementation}

\begin{figure}
  \centering
    \begin{subfigure}{0.0\linewidth}
            \phantomsubcaption
        \label{fig:pre_nms}
    \end{subfigure}
        \begin{subfigure}{0.0\linewidth}
        \phantomsubcaption
        \label{fig:nms_box}

    \end{subfigure}
    
  \begin{subfigure}{1.0\linewidth}

  \includegraphics[width=\linewidth]{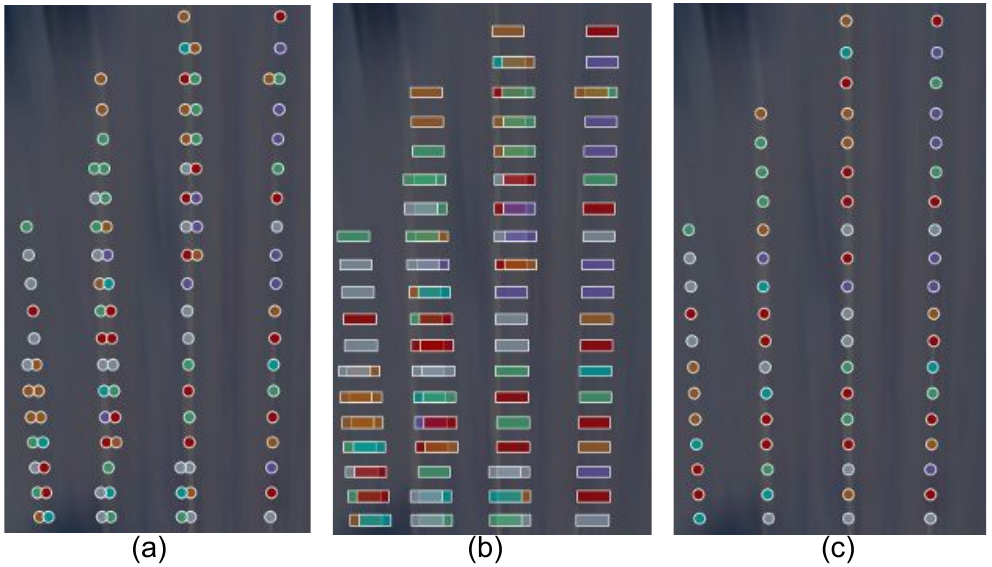}
          \phantomsubcaption

    \label{fig:post_nms}
\end{subfigure}
    \caption{PointNMS Process: (a) Initial proposals, (b) Bounding boxes centered at the proposal points, (c) Selected strongest keypoints after applying PointNMS.}
    \vspace*{-0.5cm}
\end{figure}

The PointNMS function selects only the highest-confidence keypoint from a group of proposals located within a distance of \( d_x \) from each other, with \( d_y \) being approximately zero.  

Rather than implementing a custom module for the PointNMS operation, we leverage the commonly available BoxNMS function to achieve the same result.  

The input to the function consists of the proposal locations (\( \mathbf{K}_P \)), as depicted in \cref{fig:pre_nms}. Bounding boxes are then generated around these proposals, as shown in \cref{fig:nms_box}, where the centers of the bounding boxes correspond to the positions of the proposals. The width of each box is set to \( d_x \), and the height is \( d_y \). To enhance resolution, the proposal positions are scaled using a reshaping parameter \( r \), as illustrated in \cref{fig:point_nms code}.

\section{Spatial Constraints in the Matcher}

We use the Hungarian algorithm to match predicted proposals with ground truth keypoints, as discussed in \cref{sec:keypoint_matcher}. GLane3D utilizes anchor points \( \mathbf{K}_A \) during the prediction of proposals \( \mathbf{K}_P \), where the anchor points maintain a fixed initial position that is not updated. Because of this, we apply spatial constraints to avoid matching proposals with distant ground truth keypoints.  

The first constraint concerns the lateral distance between the predicted position of a proposal, \( (\mathbf{x}_i + \Delta{x}, \mathbf{y}_i) \), and the ground truth keypoint, \( (\mathbf{x}_t, \mathbf{y}_t) \). If the lateral distance, \( |\mathbf{x}_i + \Delta{x} - \mathbf{x}_t| \), exceeds 1 meter, the match is rejected. The second constraint addresses the initial position of the anchor point, \( (\mathbf{x}_i, \mathbf{y}_i) \), and the ground truth position, \( (\mathbf{x}_t, \mathbf{y}_t) \). If the lateral distance, \( |\mathbf{x}_i - \mathbf{x}_t| \), exceeds 2 meters, the match is also rejected. The third constraint relates to the longitudinal distance between the proposal and the ground truth. Since our model does not predict an offset in the longitudinal \( y \)-axis, we reject matches if the longitudinal distance \( |\mathbf{y}_i - \mathbf{y}_t| \) exceeds 0 meters.  

We prevent matching when any of these conditions are met by replacing the corresponding value in the cost matrix with an infinite value.

\section{Graph of Keypoints to Lane Instances}

\begin{figure*}
  \centering
    \begin{subfigure}{0.0\linewidth}
            \phantomsubcaption
        \label{fig:graph}
    \end{subfigure}

  \begin{subfigure}{1.0\linewidth}

  \includegraphics[width=\linewidth]{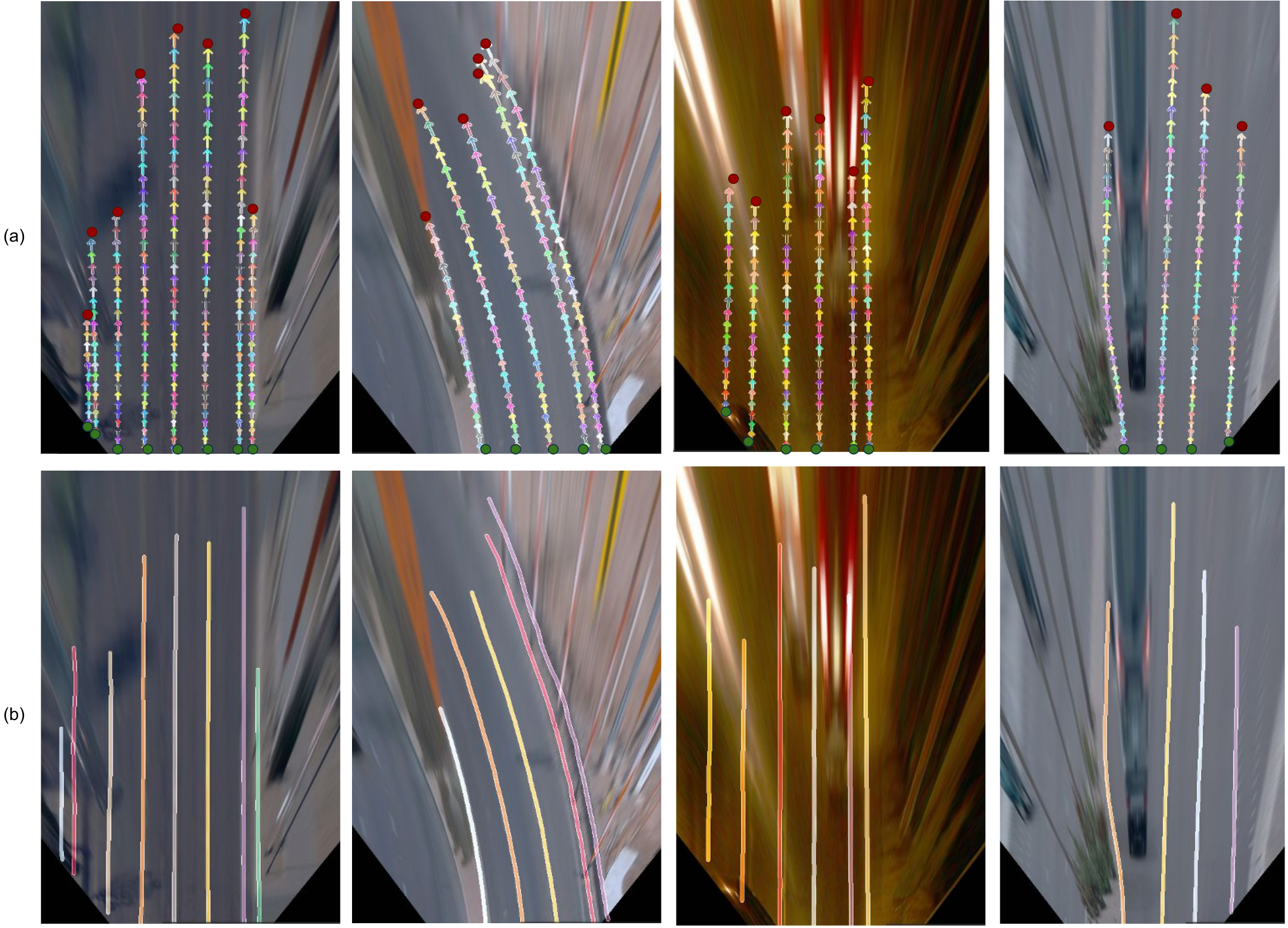}
          \phantomsubcaption
        \label{fig:instances}

\end{subfigure}
    \caption{Lane Instance Extraction: (a) Graph of keypoints, (b) Extracted lane instances.}
    \label{fig:graph_to_instances}

\end{figure*}

The output of GLane3D is a graph representing keypoints, where directed edges connect sequential keypoints, as shown in \cref{fig:graph}. Initially, we identify the start and end keypoints, which satisfy the conditions specified in \cref{formula:start_point} and \cref{formula:end_point}, respectively. These keypoints are represented as green and red points in \cref{fig:graph}. In the next step, we compute the shortest paths between the start and end keypoints using Dijkstra's algorithm. To incorporate adjacency probabilities \( \mathbf{A} \) into the shortest path estimation, we use the corresponding values from the matrix \( 1 - \mathbf{A} \), as this is a minimization problem.

\begin{table*}[b]
    \centering
    \begin{tabular}{l| c | c  c | c  c}
        \Xhline{1.3pt}
         Projection & F1-Score(\%) $\uparrow$ & \begin{tabular}{@{}c@{}}X-error \\  near(m) $\downarrow$  \end{tabular}  & \begin{tabular}{@{}c@{}}X-error \\  far(m) $\downarrow$  \end{tabular}   & \begin{tabular}{@{}c@{}}Z-error \\  near(m) $\downarrow$ \end{tabular} & \begin{tabular}{@{}c@{}}Z-error \\  far(m) $\downarrow$  \end{tabular} \\
         \hline

         IPM w/o Custom BEV & 71.6 & 0.222 & 0.275 & 0.094 & 0.125 \\
         IPM w/ Custom BEV & 72.0  & 0.239 & 0.267 & 0.093 & 0.121 \\
         LSS & 72.1 & 0.219 & 0.261 & 0.091 & 0.118 \\

         \Xhline{1.3pt}

    \end{tabular}
    \caption{Comparison of PV to BEV Projection Methods.}
    \label{tab:lss_vs_ipm}
\end{table*}

\begin{table*}
    \begin{center}
         \begin{tabular}{c | l |  l | c | c |  c |  c | c | c | c | c } 
        \Xhline{1.3pt}
         Dist. & Methods & Backbone & Sensors & All & \begin{tabular}{@{}c@{}}Up \& \\ Down  \end{tabular}   & Curve & \begin{tabular}{@{}c@{}}Extreme \\  Weather  \end{tabular}   & Night  & Inter.  &  \begin{tabular}{@{}c@{}}Merge \\  Split  \end{tabular} \\ [0.5ex] 
        \hhline{===========}
        
         \parbox[t]{2mm}{\multirow{10}{*}{\rotatebox[origin=c]{90}{\textbf{\textit{1.5m}}}}} &
         PersFormer \cite{chen2022persformer} &  EffNet-B7 & C & 50.5 & 42.4 & 55.6 & 48.6 & 46.6  & 40.0 & 50.7 \\
         & BEV-LaneDet \cite{wang2023bev}& ResNet-34 & C &  58.4 & 48.7 & 63.1 & 53.4 & 53.4  & 50.3 & 53.7 \\
         & PersFormer \cite{chen2022persformer} &  ResNet-50 & C & 53.7 & 46.4 & 57.9 & 52.9 & 47.2  & 41.6 & 51.4 \\
          & M$^2$-3DLaneNet \cite{luo2022m} &  EffNet-B7 & C+L & 55.5 & 53.4 & 60.7 & 56.2 & 51.6  & 43.8 & 51.4 \\
         &LATR \cite{luo2023latr}& ResNet-50 & C & 61.9 & 55.2 & 68.2 & 57.1 & 55.4  & 52.3 & 61.5 \\
         &LaneCPP \cite{pittner2024lanecpp}& EffNet-B7 & C & 60.3 & 53.6 & 64.4 & 56.7 & 54.9  & 52.0 & 58.7 \\
         &PVALane \cite{zheng2024pvalane}& ResNet-50 & C & 62.7 & 54.1 & 67.3 & 62.0 & 57.2  & 53.4 & 60.0 \\
          &DV-3DLane \cite{luo2024dv}& ResNet-34 & C+L & 65.4 & 60.9 & 72.1 & \underline{64.5} & 61.3  & 55.5 & 61.6 \\
         &DV-3DLane \cite{luo2024dv}& ResNet-50 & C+L & \textbf{66.8} & \underline{61.1} & 71.5 & \textbf{64.9} & \textbf{63.2}  & \textbf{58.6} & 62.8 \\
         \cline{2-11} &\mycc  Glane3D-Lite &\mycc ResNet-18 &\mycc C &\mycc 61.5 &\mycc 55.6 &\mycc 69.1 &\mycc 56.6 &\mycc 56.6 &\mycc 52.9 &\mycc 61.3\\
          \cline{2-11} &\mycc GLane3D-Base &\mycc ResNet-50 &\mycc C &\mycc 63.9 &\mycc 58.2 &\mycc 71.1 &\mycc 60.1 &\mycc 60.2 &\mycc 55.0 &\mycc 64.8 \\
        \cline{2-11}  &\mycc GLane3D-Large &\mycc Swin-B &\mycc C &\mycc 66.0 &\mycc \underline{61.1} &\mycc \underline{72.5} &\mycc 64.2 &\mycc 60.1 &\mycc \underline{58.0} &\mycc \underline{66.9} \\
       \cline{2-11}  &\mycc GLane3D-Fusion &\mycc ResNet-50 &\mycc C+L &\mycc \underline{66.6} &\mycc \textbf{61.7} &\mycc \textbf{72.7} &\mycc 63.8 &\mycc \underline{62.0} &\mycc 57.9 &\mycc \textbf{67.7} \\

        \Xhline{1.3pt}

                 \parbox[t]{2mm}{\multirow{10}{*}{\rotatebox[origin=c]{90}{\textbf{\textit{0.5m}}}}} &
         PersFormer \cite{chen2022persformer} &  EffNet-B7 & C & 36.5 & 26.8 & 36.9 & 33.9 & 34.0  & 28.5 & 37.4 \\
         & Anchor3DLane \cite{huang2023anchor3dlane}& EffNet-B3 & C & 34.9  & 28.3 & 31.8 & 30.7 & 32.2  & 29.9 & 33.9 \\
          & M$^2$-3DLaneNet \cite{luo2022m} &  EffNet-B7 & C+L & 48.2 & 40.7 & 48.2 & 49.8 & 46.2  & 38.7 & 44.2 \\
         & PersFormer \cite{chen2022persformer} &  ResNet-50 & C & 43.2& 36.3 & 42.4 & 45.4 & 39.3 & 32.9 & 41.7 \\

         &LATR \cite{luo2023latr}& ResNet-50 & C & 54.0 & 44.9 & 56.2 & 47.6 & 46.2 & 45.5  & 55.6 \\
         &DV-3DLane \cite{luo2024dv}& ResNet-34 & C+L & 63.5 & 58.6 & \underline{69.3} & 62.4 & 59.9  & \underline{53.9} & 59.3 \\
         &DV-3DLane \cite{luo2024dv}& ResNet-50 & C+L & \underline{65.2} & \underline{59.1} & 69.2 & \textbf{63.0} & \textbf{62.0}  & \textbf{56.9} & 60.5 \\
         \cline{2-11} &\mycc Glane3D-Lite &\mycc ResNet-18 &\mycc C &\mycc 53.8 &\mycc 46.7 &\mycc 57.7 &\mycc 47.9 &\mycc 47.1 &\mycc 45.8 &\mycc 55.7\\
          \cline{2-11} &\mycc GLane3D-Base &\mycc ResNet-50 &\mycc C &\mycc 57.9 &\mycc 51.0 &\mycc 61.7 &\mycc 53.5 &\mycc 53.8  &\mycc 49.4 &\mycc 60.5 \\
        \cline{2-11}  &\mycc GLane3D-Large &\mycc Swin-B &\mycc C &\mycc 61.1 &\mycc 54.2 &\mycc 64.5 &\mycc 56.8 &\mycc 55.2  &\mycc 53.6 &\mycc \underline{63.3} \\
        \cline{2-11}  &\mycc GLane3D-Fusion &\mycc ResNet-50 &\mycc C+L &\mycc \textbf{65.6} &\mycc \textbf{62.4} &\mycc \textbf{71.6} &\mycc \underline{62.9} &\mycc \underline{61.1}  &\mycc \textbf{56.9} &\mycc \textbf{66.6} \\
        \Xhline{1.3pt}
        \end{tabular}
       \caption{Extended Quantitative Results by Category on the OpenLane Dataset \cite{chen2022persformer}.}
       \label{tab:openlane_category_extend}
    \end{center}
    \vspace*{-0.3cm}

\end{table*}

\section{Custom BEV Geometry Adjustment}

As discussed in \cref{sec:special_geometry}, the anchor points \( \mathbf{K_A} \) used in Inverse Perspective Mapping (IPM) are evenly distributed across the Bird’s Eye View (BEV) space. However, when projected onto the frontal view (FV), the anchor points become sparser in regions closer to the ego vehicle and denser in areas farther away, as shown in \cref{fig:pv_a}.  

To address this sparsity near the ego vehicle, our method adjusts the distribution of anchor points \( \mathbf{K_A} \) by reducing both the longitudinal and lateral distances between keypoints as they approach the ego vehicle. Specifically, the longitudinal distances between keypoints increase linearly from 0.5 to 1.5 meters, from the ego vehicle to the farthest point, as described in \cref{formula:custom_bev_y} and \cref{formula_dy}. Meanwhile, the lateral distances between keypoints at the same longitudinal distance decrease as they approach the ego vehicle. This adjustment narrows the width of the BEV space near the ego vehicle, ensuring that the number of columns remains constant while preserving the rectangular shape of the BEV feature \( F_{BEV} \). As shown in \cref{formula:custom_bev_x_start} and \cref{formula:custom_bev_x_end}, the points in the first row begin at \( W \times \frac{1}{4} \) and end at \( W \times \frac{3}{4} \), while points in the farthest row start at 0 and end at \( W \). The lateral range increases linearly from the nearest row to the farthest row.

\begin{equation}
    \text{dy} = \frac{1}{H - 1}
      \label{formula_dy}
\end{equation}
\begin{equation}
    y_i = 0.5 + i \times \text{dy}, \quad \text{for } i = 0, 1, \dots, H - 1
  \label{formula:custom_bev_y}
\end{equation}

\begin{equation}
x_{\text{start},i} = \frac{W}{4} \left( 1 - \frac{i}{H - 1} \right)
  \label{formula:custom_bev_x_start}
\end{equation}

\begin{equation}
    x_{\text{end},i} = W \times \frac{3}{4} \left( 1 - \frac{i}{H - 1} \right) + W \times \frac{i}{H - 1}
      \label{formula:custom_bev_x_end}
\end{equation}

\section{PV to BEV: IPM vs LSS}

We compare two commonly used projection methods—Inverse Perspective Mapping (IPM) and Lift Splat Shoot (LSS)—with GLane3D. IPM projects Bird’s Eye View (BEV) locations to the frontal view using camera parameters. After projection, features extracted from corresponding locations in the frontal view are sampled, resulting in \( \mathbf{F_{BEV}} \in \mathbb{R}^{c \times H_b \times W_b} \).

Lift Splat Shoot, on the other hand, estimates depth from frontal view features \( \mathbf{F_{FV}} \in \mathbb{R}^{H' \times W' \times C} \), which is extracted from frontal view image \( \mathbf{I} \in \mathbb{R}^{3 \times H \times W} \) . Each feature vector is projected onto the corresponding BEV grid, which is calculated based on the camera position and depth estimation. Since multiple pixels may fall within the same BEV grid, LSS applies a cumulative sum trick to pool the features that fall within each grid.

A key challenge with IPM is its reliance on projections sampled from the ground surface, which can lead to inaccurate height estimations, particularly on non-flat surfaces. To compare the two projection methods, we trained our model using both LSS and IPM, with only the projection block differing between the networks, while the rest of the architecture remained the same. The results, as shown in \cref{tab:lss_vs_ipm}, indicate that there is negligible difference between the two projection methods in terms of F1-score. Since IPM requires less effort for deployment across different platforms, we opted to use IPM for our training.

\section{Extended OpenLane Results}

Category-based F1 scores with thresholds of 1.5m and 0.5m are presented in \cref{tab:openlane_category_extend}. The results indicate that camera-only GLane3D-Large achieves performance similar to that of camera + LiDAR fusion models in terms of F1 score at the 1.5m threshold. However, the differences in F1 scores between camera-only models and camera + LiDAR fusion models at the 0.5m threshold highlight the contribution of LiDAR to localization accuracy.

GLane3D-Fusion outperforms other models in most categories, as shown in \cref{tab:openlane_category_extend}. It is important to note that GLane3D-Fusion utilizes a simpler fusion approach for combining LiDAR and camera features, in contrast to the more complex fusion strategy used by DV-3DLane \cite{luo2024dv}.

\section{Extended Apollo Results}

\cref{tab:apollo_extended} shows the extended results in Apollo \cite{guo2020gen} dataset with previous methods.

\begin{table*}[hb]
     \begin{tabular}{ c | l |  l | c |  c |  c c | c c} 
    \Xhline{1.3pt}
     Subset & Methods & Backbone & F1-Score(\%)$\uparrow$ & AP(\%)$\uparrow$ &\begin{tabular}{@{}c@{}}X-error \\  near(m) $\downarrow$  \end{tabular}  & \begin{tabular}{@{}c@{}}X-error \\  far(m) $\downarrow$  \end{tabular}   & \begin{tabular}{@{}c@{}}Z-error \\  near(m) $\downarrow$ \end{tabular} & \begin{tabular}{@{}c@{}}Z-error \\  far(m) $\downarrow$  \end{tabular}  \\ [0.5ex] 
    \hhline{=========}
    
     \parbox[t]{2mm}{\multirow{6}{*}{\rotatebox[origin=c]{90}{\textit{Balanced Scene}}}} 
     & PersFormer\cite{chen2022persformer} & EffNet-B7 & 92.9 & -  & 0.054 & 0.356 & 0.010  & 0.234  \\
     & BEVLaneDet \cite{wang2023bev} & ResNet-34 & 96.9 & - & \textbf{0.016} & \textbf{0.242} & 0.020  & 0.216   \\
     & LaneCPP \cite{pittner2024lanecpp} &  EffNet-B7 & \underline{97.4} & \textbf{99.5} & 0.030 & 0.277 & 0.011  & \textbf{0.206}  \\
     & LATR \cite{luo2023latr}& ResNet-50 & 96.8 & 97.9 & 0.022 & 0.253 & \textbf{0.007}  & \textbf{0.202} \\
     & DV-3DLane \cite{luo2024dv}& ResNet-50 & 96.4 & 97.6 & 0.046 & 0.299 & 0.016  & \underline{0.213} \\
    \cline{2-8} &\mycc GLane3D (Ours) &\mycc ResNet-50 &\mycc \textbf{98.1} &\mycc \underline{98.8} &\mycc \underline{0.021} &\mycc \textbf{0.250} &\mycc \underline{0.007}  &\mycc 0.213 \\

    \Xhline{1.3pt}

    \parbox[t]{2mm}{\multirow{6}{*}{\rotatebox[origin=c]{90}{\textit{Rare Scene}}}} 
     & PersFormer\cite{chen2022persformer} & EffNet-B7 & 87.5 & -  & 0.107 & 0.782 & 0.024  & 0.602  \\
     & BEVLaneDet \cite{wang2023bev} & ResNet-34 & \underline{97.6} & - & \textbf{0.031} & \textbf{0.594} & 0.040  & 0.556   \\
     & LaneCPP \cite{pittner2024lanecpp} &  EffNet-B7 & 96.2 & \underline{98.6} & 0.073 & 0.651 & \underline{0.023}  & \underline{0.543}  \\
     & LATR \cite{luo2023latr}& ResNet-50 & 96.1 & 97.3 & 0.050 & \underline{0.600} & \textbf{0.015}  & \textbf{0.532} \\
     & DV-3DLane \cite{luo2024dv}& ResNet-50 & 95.5 & 97.2 & 0.071 & 0.664 & 0.025  & 0.568 \\
    \cline{2-8} & \mycc GLane3D (Ours) &\mycc ResNet-50 &\mycc \textbf{98.4} &\mycc \textbf{99.1} &\mycc \underline{0.044} &\mycc 0.621 &\mycc 0.023  &\mycc 0.566 \\
    
    \Xhline{1.3pt}

   \parbox[t]{2mm}{\multirow{6}{*}{\rotatebox[origin=c]{90}{\textit{Visual Variations}}}} 
     & PersFormer\cite{chen2022persformer} & EffNet-B7 & 89.6 & -  & 0.074 & 0.430 & \textbf{0.015}  & 0.266  \\
     & BEVLaneDet \cite{wang2023bev} & ResNet-34 & \underline{95.0} & - & \textbf{0.027} & \underline{0.320} & 0.031  & 0.256   \\
     & LaneCPP \cite{pittner2024lanecpp} &  EffNet-B7 & 90.4 & 93.7 & 0.054 & 0.327 & 0.020  & \textbf{0.222}  \\
     & LATR \cite{luo2023latr}& ResNet-50 & \textbf{95.1} & \textbf{96.6} & 0.045 & \textbf{0.315} & \underline{0.016}  & \underline{0.228} \\
     & DV-3DLane \cite{luo2024dv}& ResNet-50 & 91.3 & 93.4 & 0.095 & 0.417 & 0.040  & 0.320 \\
    \cline{2-8} &\mycc GLane3D (Ours) &\mycc ResNet-50 &\mycc 92.7 &\mycc \underline{94.8} &\mycc 0.046 &\mycc 0.364 &\mycc 0.020  &\mycc 0.317 \\
    \Xhline{1.3pt}

    \end{tabular}
   \caption{Extended Quantitative Results on the Apollo 3D Synthetic Dataset \cite{guo2020gen}.}
   \label{tab:apollo_extended}
\end{table*}




\end{document}